\documentclass[letterpaper, 10 pt, conference]{ieeeconf}  
\usepackage{amsmath,amsfonts}
\usepackage{algorithmic}
\usepackage{algorithm}
\usepackage{array}
\usepackage[caption=false,font=normalsize,labelfont=sf,textfont=sf]{subfig}
\usepackage{textcomp}
\usepackage{stfloats}
\usepackage{url}
\usepackage{verbatim}
\usepackage{graphicx}
\usepackage{cite}
\usepackage{cleveref}
\usepackage{booktabs}
\usepackage{multirow}
\usepackage{array}

\IEEEoverridecommandlockouts                              

\title{\LARGE \bf
A Reactive Framework for Whole-Body Motion Planning of Mobile Manipulators Combining Reinforcement Learning and SDF-Constrained Quadratic Programming
}

\author{Chenyu Zhang, Shiying Sun, Kuan Liu, Chuanbao Zhou, Xiaoguang Zhao, Min Tan, Yanlong Huang
\thanks{This paper was supported by the National Natural Science Foundation of China (No. 62203438 and 62103410), in part by the Science and Technology Project of Beijing (No. Z231100007123008)}
\thanks{Chenyu Zhang, Shiying Sun, Kuan Liu, Chuanbao Zhou, Xiaoguang Zhao and Min Tan are with State Key Laboratory of Multimodal Artificial Intelligence Systems, Institute of Automation, Chinese Academy of Sciences, Beijing
100190, China and also with School of Artificial Intelligence, University of Chinese Academy of
Sciences, Beijing 100049, China {\tt\small zhangchenyu2020@ia.ac.cn sunshiying2013@ia.ac.cn liukuan2023@ia.ac.cn zhouchuanbao2021@ia.ac.cn min.tan@ia.ac.cn xiaoguang.zhao@ia.ac.cn}}%
\thanks{Yanlong Huang is School of Computing, University of Leeds, Leeds LS29JT, UK. y.l.{\tt\small huang@leeds.ac.uk }}%
}

\begin{document}

\maketitle

\thispagestyle{empty}
\pagestyle{empty}

\begin{abstract}
As an important branch of embodied artificial intelligence,
mobile manipulators are increasingly applied in intelligent services, but their redundant degrees of freedom also limit efficient motion planning in cluttered environments. 
To address this issue, this paper proposes a hybrid learning and optimization framework for reactive whole-body motion planning of mobile manipulators.
We develop the Bayesian distributional soft actor-critic (Bayes-DSAC) algorithm to improve the quality of value estimation and the convergence performance of the learning. 
Additionally, we introduce a quadratic programming method constrained by the signed distance field to enhance the safety of the obstacle avoidance motion. 
We conduct experiments and make comparison with standard benchmark.
The experimental results verify that our proposed framework significantly improves the efficiency of reactive whole-body motion planning, reduces the planning time, and improves the success rate of motion planning. 
Additionally, the proposed reinforcement learning method ensures a rapid learning process in the whole-body planning task.
The novel framework allows mobile manipulators to adapt to complex environments more safely and efficiently.
\end{abstract}

\section{Introduction}
Mobile manipulators represent a significant branch in the field of embodied artificial intelligence\cite{duan2022survey}, and
have demonstrated significant potential in a variety of interactive tasks, such as industrial assembly, manufacturing, intelligent services and medical assistance, due to their mobility and manipulability. 
These robots need to perform obstacle avoidance navigation and manipulation in large, complex environments such as factories and offices.

The redundant degrees of freedom of mobile manipulator affect end-effector(EE)'s operations due to mobile base movements.
Despite increased null space flexibility, motion planning of the robot remains a significant challenge.
Early control schemes\cite{ZHANG2024104644} consider the motion of mobile base and manipulator separately, this control strategy substantially constricts the solution space, challenging in complex scenarios.
The current focus is on whole-body motion planning methods. 
While reinforcement learning based coordinated control methods\cite{Kindle2020} are capable of extracting environmental features to generate velocities, the planning results often violate the dynamic constraints.
Although optimization-based control methods\cite{pankertPerceptiveModelPredictive2020a} achieve coordinated control by adding constraints, they are limited by the constraints' expressive capabilities and environmental understanding. 
Additionally, traditional velocity control methods only consider deviations relative to the target, the planning results are likely to fall into local minima. 
Currently, existing planning methods are generally incapable of achieving efficient obstacle avoidance\cite{Liu2021}.

In order to handle the challenges mentioned above, we propose a novel
hybrid learning and optimization framework for reactive whole-body motion planning of mobile manipulators. 
The main contributions of the paper are:

(1) A hybrid learning and optimization framework for reactive whole-body motion planning of mobile manipulator. 
It uses reinforcement learning at the high planning layer to perceive the surrounding environment  and suggest end-effector velocity planning for mobile manipulators; the lower-level control strategy employs quadratic programming to solve joint velocities, ensuring reliable, swift, and elegant robot control.

(2) An off-policy reinforcement learning method named Bayesian distributional soft actor-critic
(Bayes-DSAC) for the mobile manipulator whole-body planning task.
The algorithm adopts a Bayesian fusion approach to update the target value function distribution, which avoids the excessive underestimation of the value estimation. 
This improvement has enhanced the quality of the value function estimation, allowing for a higher replay-ratio, which improves the sample efficiency and convergence speed of the algorithm.

(3) A signed distance field(SDF) constrained QP for the control layer.
It leverages parallel queries of the SDF to obtain the distances and compute a repulsive vector field. It is used to constrain the joint velocities in QP for avoiding collisions, thus enhancing the safety of robot motion.

(4) Extensive experimental validation of the proposed hybrid framework.
Simulation results verify that our framework can provide higher-quality whole-body planning results compared to traditional strategies.
Moreover, the novel RL method achieves better average returns to the whole-body planning task.

The rest of the paper is organized as follows. Section II
introduces the related work. Section III details the methodology for the  
hybrid motion planning framework. 
The experimental results and the discussions are presented in Section IV. Finally, Section V concludes with remarks and future directions.
\section{RELATED WORK}
\subsection{Motion planning of mobile manipulator}
Motion planning of mobile manipulators is a complex problem while satisfying collision avoidance, joint constraints, and velocity smoothness.
Traditional solutions generally decomposed the challenge into sequential stages,
first positioning the base and then planning manipulator movements, leading to poor operational efficiency\cite{stibingerMobileManipulatorAutonomous2021a}. 
Therefore, this paper concentrates on whole-body collaborative motion planning for mobile manipulators.

Optimization-based methods, such as Model Predictive Control (MPC)\cite{Chiu2022}, improved robustness through forward rolling optimization, but they required 
precision mode and lack real-time capability. Although perceptive MPC(P-MPC)\cite{pankertPerceptiveModelPredictive2020a} enhanced obstacle avoidance, they were limited by the optimization speed and accuracy of environmental perception. Reactive planning methods, such as the NEO method\cite{havilandNEONovelExpeditious2021a}, offered fast response speed. 
Haviland\cite{Holistic} proposed a reactive velocity planning method to improve speed based on QP, but they did not consider obstacle avoidance and local optimally issues.
RAMPAGE\cite{yangRAMPAGEWholeBodyRealTime2024} integrated perception and control to achieve real-time planning through Euclidean Signed Distance Fields(ESDF) map search. However, the adaptability and safety need to be further improved due to perception limitations.
Wang et al.~\cite{wangReactiveMobileManipulation2024} proposed a planning strategy for dual-trajectory tracking. However, this method requires a global map and may lead to repetitive planning.

Reinforcement learning methods\cite{Kindle2020} performed online obstacle avoidance through feature extraction, but they were limited by discrete action spaces.
Jauhri et al.~\cite{Jauhri2022} proposed mixed action space reinforcement learning by Gumbel-Softmax, but obstacle avoidance planning is still influenced by the environment. 
Honerkamp et al.~\cite{Honerkamp2022} decomposed the task space and combined end-effector trajectory generation with mobile base navigation to compress the parameter space, but collision risk remains.
Combining reinforcement learning with optimization methods, such as Model-based Reinforcement Learning (MBRL)\cite{Hoeller2020,Brito2021}, strengthened policy through the kinematic model to improve disturbance resistance.
Yao et al.~\cite{Yao2022} proposed a control method based on disturbance prediction, combining forward prediction and reinforcement learning to achieve stable control.

In summary, optimization methods excelled in accuracy and model utilization, while reinforcement learning methods perform better in environmental understanding.
This paper will leverage the strengths of both approaches to achieve fast and accurate reactive planning.

\subsection{Off-policy reinforcement learning}
The whole-body planning of mobile manipulator robots is a complex issue requiring long-term strategies in a high-dimensional continuous action space.
Off-policy reinforcement learning becomes a crucial branch due to its efficient data utilization and convergence of the algorithms. 
For continuous action spaces, algorithms like TD3\cite{pmlr-v80-fujimoto18a} realized continuous control through actor-critic architecture. 
Moreover, SAC\cite{haarnojaSoftActorCriticOffPolicy2018} employed an increase in 
expected policy entropy, thereby improving the efficiency of environmental exploration, with more stable algorithm performance. 
To control the overestimation bias of the policy function, methods such as TQC\cite{kuznetsovControllingOverestimationBias} used combined ensemble strategies, controlling the value bias with hyperparameters. REDQ\cite{chenRandomizedEnsembledDouble2021} reduced value estimation bias by a random ensemble of multiple Q networks to handle environmental uncertainties well. 
Based on these, DSAC-T\cite{duanDSACTDistributionalSoft2023a} adopted return distribution estimation, which improved the accuracy of the value function estimation, mitigated learning instability. DSAC-T effectively handled the long-term control problems, making it one of the best-performing algorithms for whole-body motion planning task.
However, these methods typically employed clipped double Q-learning, which led to a significant underestimation of the return function in learning. Bayesian controller fusion(BCF)\cite{Arcari2023} could integrate two distribution functions without overestimation and underestimation.

Therefore, this paper adopts a Bayesian controller fusion based on DSAC-T to estimate the value function, aiming to further improve the planning quality and convergence speed of reinforcement learning algorithms in whole-body planning problems.
\begin{figure*}[htbp]
        \centering
        \includegraphics[width=\linewidth,scale=0.3]{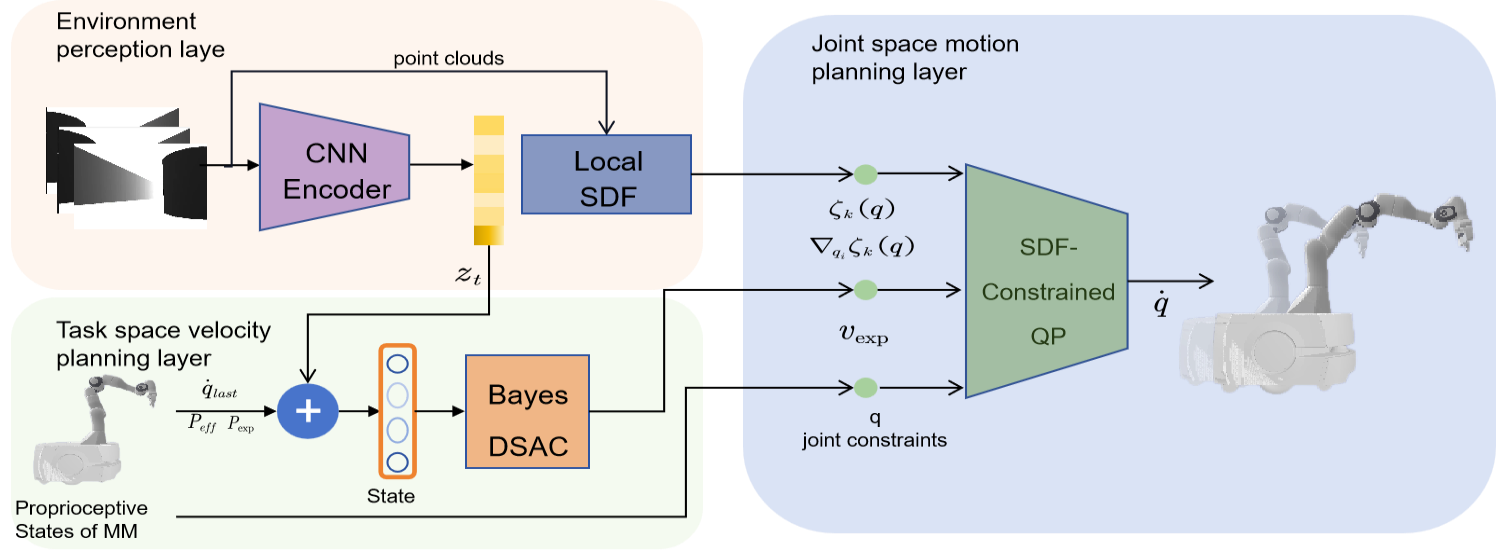}
        \caption{Overview of 
        the hybrid learning and optimization framework for reactive whole-body motion planning in mobile manipulators.}
        \label{fig:process}
      \end{figure*}
\section{METHODOLOGY}
An overview of proposed hybrid learning and optimization framework in this paper is shown in Fig~\ref{fig:process}.
It consists of three parts:
(1) Environment perception layer, which uses depth images to provides the velocity planning layer with features of obstacles.
Additionally, we establish a local SDF map of the current point cloud to provide joint restrictions.
(2) Task space velocity planning layer, which uses environment features and proprioceptive states to calculate the desired end-effector(EE) velocity via Bayes-DSAC. 
This layer provides a task space guidance for whole-body control.
(3) Joint space motion planning layer, which achieve mobile manipulator whole-body control by solving a QP problem. 
During the optimization process, an intentional error is allowed for the velocity to minimize costs while meeting the constraints. 
Additionally, by querying the SDF map and calculating gradients, we can impose velocity constraints on potential collision joints.

\subsection{Environment perception layer}
The environment perception module is divided into two parts. One part overlaps the depth images of the previous 5 frames and inputs them into the CNN-based environment encoder module to obtain a 32-length environment feature $z_t$. The environment perception module is jointly trained during the reinforcement learning process. 
Another part converts the previous 5 frames into point clouds according to the position and establishes a local map. 
Unlike querying the nearest distance by the environment ESDF using rectangles in other motion planning methods\cite{8758904}, 
the non-convex structure of the mobile manipulator makes it difficult to query the nearest distance once, necessitating the use of cylinders to approximate its structure. It reduces the accuracy of collision detection.
Consequently, our framework employs a parallel robot-centric SDF, transforming the query objects into point clouds, to ensure both speed and precision.
We use SDF to query the distances $\zeta_k(q)$ to the $k$ link under the current joint configuration $q$ and calculate the distance gradients $\nabla_{q_i}\zeta_k(q)$  for each joint $q_i$.

\subsection{Bayes-DSAC for task space velocity planning layer}
The mobile manipulator EE reaching task is a complex issue that requires long-term, high-precision planning strategies, which can be represented as Markov decision process with continuous state space $S$ and continuous action spaces $A$.
The agent get reward $r\left(s_t, a_t\right)$, and the state transition probability is $p\left(s_{t+1}\mid s_t, a_t\right)$.
The agent's strategy is defined as $\pi(a_t\mid s_t)$, which represents the probability distribution of actions taken in a given state.
The $\rho_\pi\left(s_t\right)$ and $\rho_\pi\left(s_t,a_t\right)$ denote the state and state-action distributions induced by the policy.
The objective of RL is to learn a policy that maximizes the expected future accumulated return.
Here, we adopt the maximum entropy objective\cite{haarnojaSoftActorCriticOffPolicy2018}, which augments the reward $r$ with policy entropy.
According to the Distributional Soft Policy Iteration (DSPI) framework\cite{duanDistributionalSoftActorCritic2022}, we define the soft state-action return in a distributional form:
\begin{equation}
  \begin{split}
  Z^\pi\left(s_t, a_t\right) =&r_t+\gamma \sum_{i=t+1}^{\infty} \gamma^{i-t-1}\left[r_{i}-\alpha \log \pi\left(a_{i} \mid s_{i}\right)\right]\\
  &Q^\pi(s_t, a_t)=\mathbb{E}\left[Z^\pi(s_t, a_t)\right]
\end{split}
\end{equation}
where $Z^\pi$ is a random variable and $\gamma\in(0,1)$ is the discount factor.
Correspondingly, for a policy $\pi$, it learns the soft Q-value by the distributional version of policy iteration:
\begin{equation}
  \begin{split}
  \mathcal{T}_{\mathcal{D}}^\pi Z(s_t, a_t) \stackrel{D}{=} r+\gamma\left(Z\left({s}_{t+1}, {a}_{t+1}\right)-\alpha \log \pi\left({a}\mid {s}_{t+1}\right)\right)
\end{split}
\end{equation}
where $\mathcal{T}^\pi$ is the soft Bellman operator and $\stackrel{D}{=}$ represents that two random variables have the same probability distribution. The $\mathcal{Z}$ represents the distribution of soft state-action return $Z^\pi$.
In policy iteration, it updates the $\mathcal{Z}$ by minimizing the Kullback-Leibler (KL) divergence between the $\mathcal{T}_{\mathcal{D}}^\pi \mathcal{Z}_\text {target}$ and $\mathcal{Z}$.
During the policy evaluation process, it updates the Critic parameters $\theta$ by minimizing this:
\begin{equation}
J_{\mathcal{Z}}(\theta)=-\underset{\mathcal{B},Z}
{\mathbb{E}}\left[\log \mathcal{P}\left(\mathcal{T}_{\mathcal{D}}^{\pi_{\phi^{\prime}}}Z(s,a) \mid \mathcal{Z}_\theta(\cdot \mid s_t, a_t)\right)\right]
\end{equation}
where $\phi^{\prime}$ and $\theta^{\prime}$ are the parameters of target critic and actor networks, while $\mathcal{B}$ denotes the buffer. Here we model $\mathcal{Z}$ as a Gaussian distribution, which can be represented as $Z_\theta(s_t, a_t) \sim \mathcal{N}(\mu_\theta(s_t, a_t), \sigma_\theta(s_t, a_t)^2)$. Therefore, $Q_\theta(s_t, a_t)=\mu_\theta(s_t, a_t)$ is the mean of value distribution and the update gradient of value network is\cite{duanDSACTDistributionalSoft2023a}:
\begin{equation}
  \begin{split}
&\nabla_\theta J_{\mathcal{Z}}(\theta) \approx \mathbb{E}\left[-\frac{\left(\mathcal{T}_{\mathcal{D}}^{\pi_{\phi^{\prime}}}Z(s_t,a_t)-Q_\theta(s_t, a_t)\right)}{\sigma_\theta(s_t, a_t)^2} \nabla Q_\theta(s_t, a_t)\right.\\
&-\left.\frac{\left(\overline{\mathcal{T}_{\mathcal{D}}^{\pi_{\phi^{\prime}}}Z(s_t,a_t)}-Q_\theta(s_t, a_t)\right)^2-\sigma_\theta(s_t, a_t)^2}{\sigma_\theta(s_t, a_t)^3} \nabla \sigma_\theta(s_t, a_t)\right]\\
&\quad \quad \overline{\mathcal{T}_{\mathcal{D}}^{\pi_{\phi^{\prime}}}Z(s_t,a_t)}=Clip(\mathcal{T}_{\mathcal{D}}^{\pi_{\phi^{\prime}}}Z(s_t,a_t),\\
&\qquad \qquad \qquad Q_\theta(s_t, a_t)-b, Q_\theta(s_t, a_t)+b)
\label{eq6}
\end{split}
\end{equation} 
where $b$ is the boundary of the constraint. 
To avoid gradient explosion caused by the excessive square of TD error, here the $\mathcal{T}_{\mathcal{D}} ^{\pi_{\phi^{\prime}}} Z(s_t, a_t)$ in the variance-related gradient is constrained to improve the stability of learning. 
To reduce the instability of mean-related gradient updates caused by the random TD error return, it uses the expectation of 
$\mathcal{T}_{\mathcal{D }}^{\pi_{\phi^{ \prime}}}Z(s_t, a_t)$ to replace the random target return:
\begin{equation}
    T_q =r+\gamma\left(Q_{\theta^{ \prime}}\left(s_{t+1}, a_{t+1}\right)-\alpha \log \pi_{\phi^{ \prime}}\left({a}_{t+1}\mid {s}_{t+1}\right)\right)
  \label{eq7}
\end{equation} 

It employs twin value distribution learning, which trains $\theta_1,\theta_2$ separately, and selects the value distribution with the smaller $Q_\theta(s_t, a_t)$ to calculate the target return in critic update.
This method can effectively reduce the overestimation of the target return distribution, but this direct minimization approach can lead to an excessive underestimation of the distribution. 
And it results in excessively conservative strategies, making the policy optimization converge more slowly. This becomes even more apparent in tasks involving whole-body motion planning in high-dimensional continuous action spaces.

\begin{figure}[htp]
  \centering
  \includegraphics[scale=0.4]{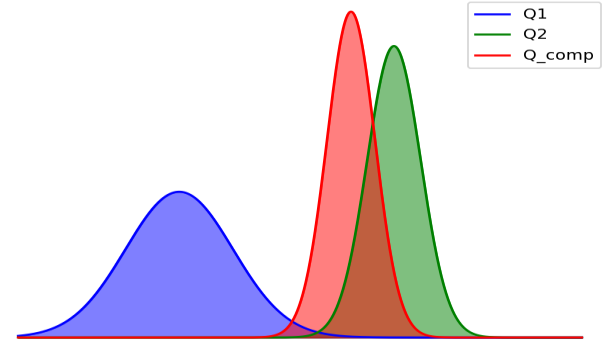}
  \caption{The result of estimation the ensemble of critics by BCF.
  The composite Q distribution avoids underestimation and mitigates the overestimation of value estimation.}
  \label{fig:Bayes}
\end{figure}

Here, we propose the Bayesian  Distributional Soft Actor Critic(Bayes-DSAC) approach.
Unlike REDQ and TQC, which use 
bias control hyperparameters to control the value approximation bias, we use BCF to compute the ensemble of critics, shown in Fig~\ref{fig:Bayes}.
It avoids underestimation and mitigates overestimation without increasing hyperparameters.
In the algorithm, we actually use the two independent distributions $\mathcal{Z}_{\theta_1},\mathcal{Z}_{\theta_2}$ obtained from fitting networks to estimate the soft state-action return $Z$. And we adopt the likelihood estimation to extend the Bayesian formula.
\begin{equation}
  \begin{split}
  p(Z \mid \mathcal{Z}_{\theta_1},&\mathcal{Z}_{\theta_2})=\frac{p\left(\mathcal{Z}_{\theta_1}, \mathcal{Z}_{\theta_2} \mid Z\right) p(Z)}{p\left(\mathcal{Z}_{\theta_1}, \mathcal{Z}_{\theta_2}\right)}\\
  &=\frac{p\left(Z\mid \mathcal{Z}_{\theta_1}\right)p\left(Z\mid \mathcal{Z}_{\theta_2}\right)p(\mathcal{Z}_{\theta_1})p(\mathcal{Z}_{\theta_2})}{p(Z)p\left(\mathcal{Z}_{\theta_1}, \mathcal{Z}_{\theta_2}\right)}
\end{split}
\end{equation}

In this way, we fuse the approximate value estimations to get the hybrid soft state-action return distribution $\mathcal{Z}_\theta^ {hyb} \sim \mathcal{N}(Q_\theta^ {hyb}, \sigma_\theta^ {hyb})$\cite{Arcari2023}.
\begin{equation}
  \begin{split}
    Q_\theta^ {hyb}&=\frac{Q_{\theta_1}\sigma_{\theta_2}^2+Q_{\theta_2} \sigma_{\theta_1}^2}{\sigma_{\theta_1}^2+\sigma_{\theta_2}^2} \\
  \sigma_\theta^ {hyb^2}&=\frac{\sigma_{\theta_1}^2 \sigma_{\theta_2}^2}{\sigma_{\theta_1}^2+\sigma_{\theta_2}^2},
\label{eq11}
\end{split}
\end{equation}

Note that, we have not considered the constant related to the uniform prior.
According to \eqref{eq6} and \eqref{eq7}, we obtain the critic gradient in policy evaluation:
\begin{equation}
  \begin{split}
&\nabla_{\theta_i} J_{\mathcal{Z}}({\theta_i})\approx \underset{Z_{hyb}\sim \mathcal{Z}_{hyb}}
{\mathbb{E}}\left[-\frac{\left(T_{q}^{hyb}-Q_{\theta_i}(s_t, a_t)\right)}{\sigma_{\theta_i}(s_t, a_t)^2} \nabla_{\theta_i} Q_{\theta_i}\right.\\
&-\left.\frac{\left(\overline{T_{z}^{hyb}}-Q_{\theta_i}(s_t, a_t)\right)^2-\sigma_{\theta_i}(s_t, a_t)^2}{\sigma_{\theta_i}(s_t, a_t)^3} \nabla_{\theta_i} \sigma_{\theta_i}\right]\\
&T_{q}^{hyb}=r+\gamma\left(Q_{{\theta^{\prime}}}^{hyb}\left(s_{t+1}, a_{t+1}\right)-\alpha \log \pi_{\phi^{ \prime}}\left({a}_{t+1}\mid {s}_{t+1}\right)\right)\\
&\qquad \qquad T_{z}^{hyb}=\mathcal{T}_{\mathcal{D }}^{\pi_{\phi^{ \prime}}}Z(s_t, a_t) \mid_{Z(s_t,a_t)\sim\mathcal{Z}_{\theta^{\prime}}^{hyb}}
\end{split}
\end{equation} 
where $\mathcal{Z}_{\theta^{\prime}}^{hyb}$ is the hybrid distribution calculated from the target network parameters$\theta^{\prime}_1,\theta^{\prime}_2$.

During the policy improvement process, the actor is improved by maximizing the entropy-argumented soft state-action return.
Similarly, the objective of the actor is rephrased by the twin value distribution and Bayesian controller fusion.
\begin{equation}
  J_\pi(\phi)={\mathbb{E}}\left[Q_\theta^ {hyb}(s_t, a_t)-\alpha \log \left(\pi_\phi(a_t \mid s_t)\right)\right]
\end{equation}
where $\alpha$ is a temperature parameter to balance the exploitation and exploration.
The Bayes-DSAC converges more quickly during training and can adopt a higher Replay-Ratio for training.

In the reactive whole-body motion planning framework for mobile manipulator, we use Bayes-DSAC to calculate the desired EE velocity through environment features and proprioceptive states. Here, the joint states, camera-perceived depth-images and positional deviation between the EE and the target are used as the state $s_t=(s_t ^ {state},s_t ^ {depth},s_t ^ {dev})$, with the velocity of the EE as the action $a=(v_x, v_y, v_z, \omega_x, \omega_y, \omega_z)$.
The reward function is designed as:
\begin{equation}
\begin{aligned}
	&r_{dev}=-norm\left( s_{t}^{dev} \right) ,\,\,\,\,\,\,\ r_{follow}=-norm\left( v_{dir}-a \right) ,\\
	&r_{goal}=25\cdot 1_{\text{norm}\left( s_{t}^{dev} \right) <\delta},\,\,\ r_{time}=-t_{run},\\
	&v_{dir}=p_{serv}\left( s_{t}^{dev},s_{t}^{state} \right) ,\ r_{coll}=\left\{ \begin{matrix}
	0&		y>s_d\\
	log\left( y \right)&		0<y<s_d\\
	-25&		y<0\\
\end{matrix} \right. ,\\
&r_{total}=\lambda _{\text{dev}}r_{\text{dev}}+\lambda _{\text{fol}}r_{\text{follow}}+r_{\text{time}}+r_{\text{goal}}+r_{\text{coll}}\\
\end{aligned}
\label{eq14}
\end{equation}
where $\lambda_{\mathrm{dev}}, \lambda_{\mathrm{fol}}$ are constants to balance the objective. $v_{dir }$ is the directly calculated position-based velocity.
The trained high-level layer employs Bayes-DSAC to learn collision-free reachable planning. This provides a foundation for the joint space motion planning layer to calculate and generate continuous, high-quality whole-body control results.

\subsection{SDF-constrained QP for joint space control layer}
To ensure that the mobile manipulator can collision-free and safely execute the desired velocity from high level planning layer,
we extend the reactive holistic planner\cite{Holistic} with the SDF constrained.
The controller adopts the QP to convert the EE speed into joints velocity while employing a slack vector to better meet constraints and achieve improved control effects.
The QP framework based on reactive safety control and EE velocity tracking is structured as follows:
\begin{equation}
  \begin{aligned}
  \min _x \qquad  &\frac{1}{2} x^{\top} \mathcal{Q} x+\mathcal{C}^{\top} x \\
  \text { s.t. }\qquad  &\left(\mathcal{J}_e(q) \quad \mathcal{I}_{6\times6} \right) x =\nu_e \\
  &\mathcal{A}_{comb} x  \leq \mathcal{B}_{comb} \\
  &\mathcal{X} ^-  \leq x \leq \mathcal{X} ^+
  \end{aligned}
  \end{equation}
where the $x$ is the decision variable composed of the joint velocities 
$\dot{q}=\left\{ \dot{q}^b \in \mathbb{R}^{2},\dot{q}^a \in \mathbb{R}^{7}\right\}$ 
and the slack vector $\delta^{v} \in \mathbb{R}^{6}$. And $\mathcal{J}_e(q)$ is the manipulator jacobian of the joint $q$.
The objective fuction of the optimization is to minimize the norm of joint velocities while maximizing the manipulability of the mobile manipulator.
The weight and term is defined by
\begin{equation}
  \mathcal{Q}=\left(\begin{array}{cc}
    \operatorname{dg}\left(\lambda_{\dot{q}}\right) & \mathbf{0}_{9 \times 6} \\
    \mathbf{0}_{6 \times 9} & \operatorname{dg}\left(\lambda_\delta\right)
    \end{array}\right) \ \
    \mathcal{C}=\left(\begin{array}{c}
        -k_\epsilon \theta_\epsilon\\
        0\\
        -\mathbf{J}_m\left(q_a\right)\\
        \mathbf{0}_{6 \times 1}
      \end{array}\right)
\end{equation}
where $\lambda_{\dot{q}}=\left\{\frac{1}{s_t ^ {dev}} \quad s_t ^ {dev} \right\}$ and $\lambda_\delta$ are weights to constrain joint velocity and adjust the slack norm. 
The term $\mathcal{C}$ is to maximaize the manipulability $\mathbf{J}_m\left(q_a\right)$ of the manipulator and minimize the manipulator angle $\theta_\epsilon=\text{atan2}({}^bT_{e,y},{}^bT_{e,x})$ \cite{Holistic}.

We employ the inequality constraints to ensure that the joint are prevented from reaching extreme and singular positions as well as avoiding collisions with the environment.
The velocity dampers$\mathcal{A}_{damper},\mathcal{B}_{damper}$ is used to constrain the joint positions.
\begin{equation}
  \begin{aligned}
  \mathcal{A}_{damper} =\left(\mathbf{1}_{9 \times 15}\right) \ \ \mathcal{B}_{damper} =\left(\begin{array}{c}
  \mathbf{0}_{2 \times 1} \\
  \eta \frac{\rho_0-\rho_s}{\rho_i-\rho_s} \\
  \vdots \\
  \eta \frac{\rho_7-\rho_s}{\rho_i-\rho_s}
  \end{array}\right)
\end{aligned}
\end{equation}
where $\rho_{0-7}$ is the distance to the nearest joint limit, $\rho_i$ and $\rho_s$ are the influence and minimum distance for a joint.

Previous environmental constraint methods\cite{havilandNEONovelExpeditious2021a,yangRAMPAGEWholeBodyRealTime2024} typically sample points on mobile manipulator, then they query the nearest distance in the SDF map of environment and compute the gradient. Subsequently, joint velocities were constrained through the Jacobian matrix and velocity damper. This process reduce the precision of collision detection, particularly in non-convex environments. Furthermore, each sampled point corresponds to an inequality constraint, resulting in overall lower computational accuracy and efficiency.
We propose a new obstacles avoidance constant, which uses the robot-centric SDF to force the robot's joints away from the collision boundaries. It queries the nearest distances between the point clouds and each link, and compute the gradient with respect to joint variations. This method precludes sampling of the robot's surface, thereby enhancing the accuracy of collision detection. Moreover, by directly calculating and constraining velocities in the joint space, we reduce the number of constraint equations, thus improving computational efficiency.

The obstacles avoidance constant is expressed as:
\begin{equation}
  \begin{aligned}
  &\mathcal{A}_{avoi} =\left(\begin{array}{lc}
    -\nabla_{q_1}\zeta_1(q) \cdots-\nabla_{q_9}\zeta_1(q) &\quad\\
    \qquad\vdots \qquad\cdots\quad\qquad\vdots \quad &\mathbf{0}_{9\times6}\\
    -\nabla_{q_1}\zeta_9(q) \cdots-\nabla_{q_9}\zeta_9(q) \qquad &\quad\\
    \end{array}\right)\in \mathbb{R}^{9\times15}  \\
  &\mathcal{B}_{avoi} =\left(\begin{array}{c}
  \zeta_1(q)-\psi\\
  \vdots\\
  \zeta_9(q)-\psi\\
  \end{array}\right) \in \mathbb{R}^9\\
\end{aligned}
\end{equation}
where $\psi$ is the desired minumum distance to the obstacle.
$\nabla_{q_i}\zeta_k(q)$ describes how joint $i$ affects the distance between link k and the obstacle, forming a repulsive vector field in the joint space.
By relaxing the constraint, when link k approaches an obstacle, $\zeta_k(q)-\psi$ becomes negative, thereby constraining the joint velocity to align with the direction that moves the dangerous link away from the obstacle. This allows the robot to satisfy collision avoidance constraint.
The inequality constraint is expressed as:
\begin{equation}
  \mathcal{A}_{comb}=\left(\begin{array}{c}\mathcal{A}_{damper}\\\mathcal{A}_{avoi}\end{array}\right) \ \mathcal{B}_{comb}=\left(\begin{array}{c}\mathcal{B}_{damper}\\\mathcal{B}_{avoi}\end{array}\right)
\end{equation}
The SDF-constrained QP for control layer collaborates the motion of the mobile base and manipulator,translating the EE velocity from task space planning layer into a reactive joint velocity planning result, ensuring the robot avoids collisions and completes its movement tasks.

\section{experimental results and discussions}
\subsection{Bayes-DSAC Training Validation}
In order to verify the effectiveness of our proposed Bayes-DSAC in the whole-body motion planning task, we use several online and off-policy RL methods, such as SAC, TD3, REDQ, TQC, DSAC-T and PPO as baselines during the training process. 
\begin{figure}[htp]
  \centering
  \includegraphics[width=\linewidth,scale=0.8]{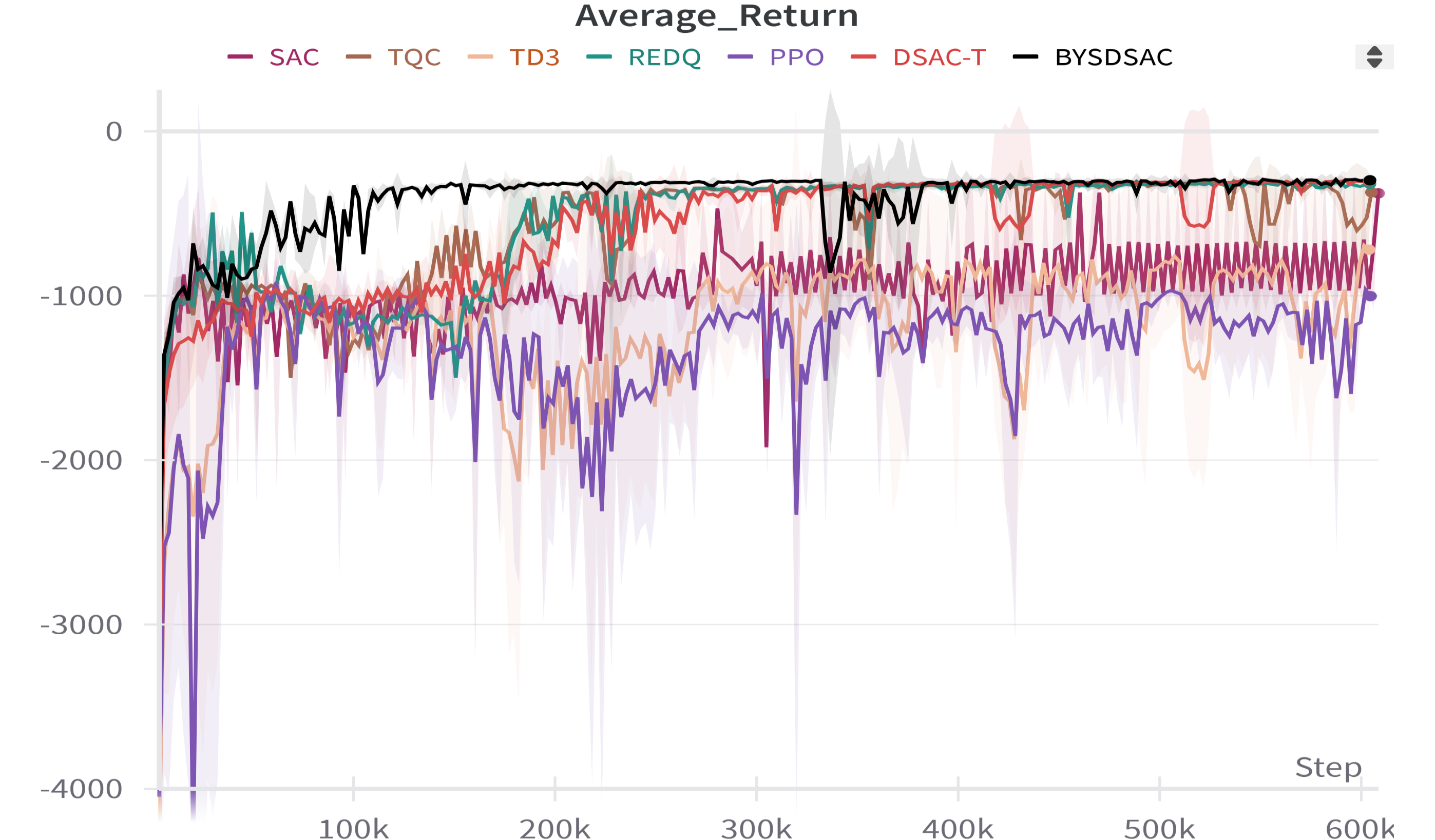}
  \caption{Training curves on benchmarks.The solid line represents the average of five runs, and the shaded area indicates the 95\% confidence interval. From the results, it can be observed that Bayes-DSAC achieves the highest average return and converges the fastest.}
  \label{fig:result1}
\end{figure}

During training, RL methods are used to generate the desired task-space EE velocities.
The same network structure and training parameters are used for several sets of algorithms, and the actor and critic share an image feature extraction network. SAC uses a Double-Q learning framework; REDQ computes 10 Q-functions simultaneously and randomly selects two for each update, using the minimum value for the update. It uses the replay-ratio of 20.
We use different random seeds to run each algorithm five times independently, with evalutions every 6000 iterations.
The maximum length of each episode is 1000 timesteps and the corresponding average return is calculated over 2 episodes without action noise.
The training results and curves are shown in Fig~\ref{fig:result1}.
The results of rapid convergence and the highest average reward indicate that the proposed Bayes-DSAC matches and surpasses all other baselines in terms of performance in whole-body motion planning tasks.
From the perspective of final average return, Bayes-DSAC can achieve results similar to DSAC-T, with some improvements over PPO, SAC, TQC, etc. 
This shows that using value distribution learning can significantly enhance the algorithm's capability, as it provides a more accurate estimation of the value, thereby mitigating some overestimation bias. 
However, selecting the minimal q-function employed by DSAC-T leads to a slight underestimation of the value. 
Our approach utilizes Bayesian estimation to mitigate this issue, resulting in an accelerated rate of reward convergence, and it also avoids overestimation bias. 

\begin{figure}[htp]
  \centering
  \includegraphics[width=\linewidth]{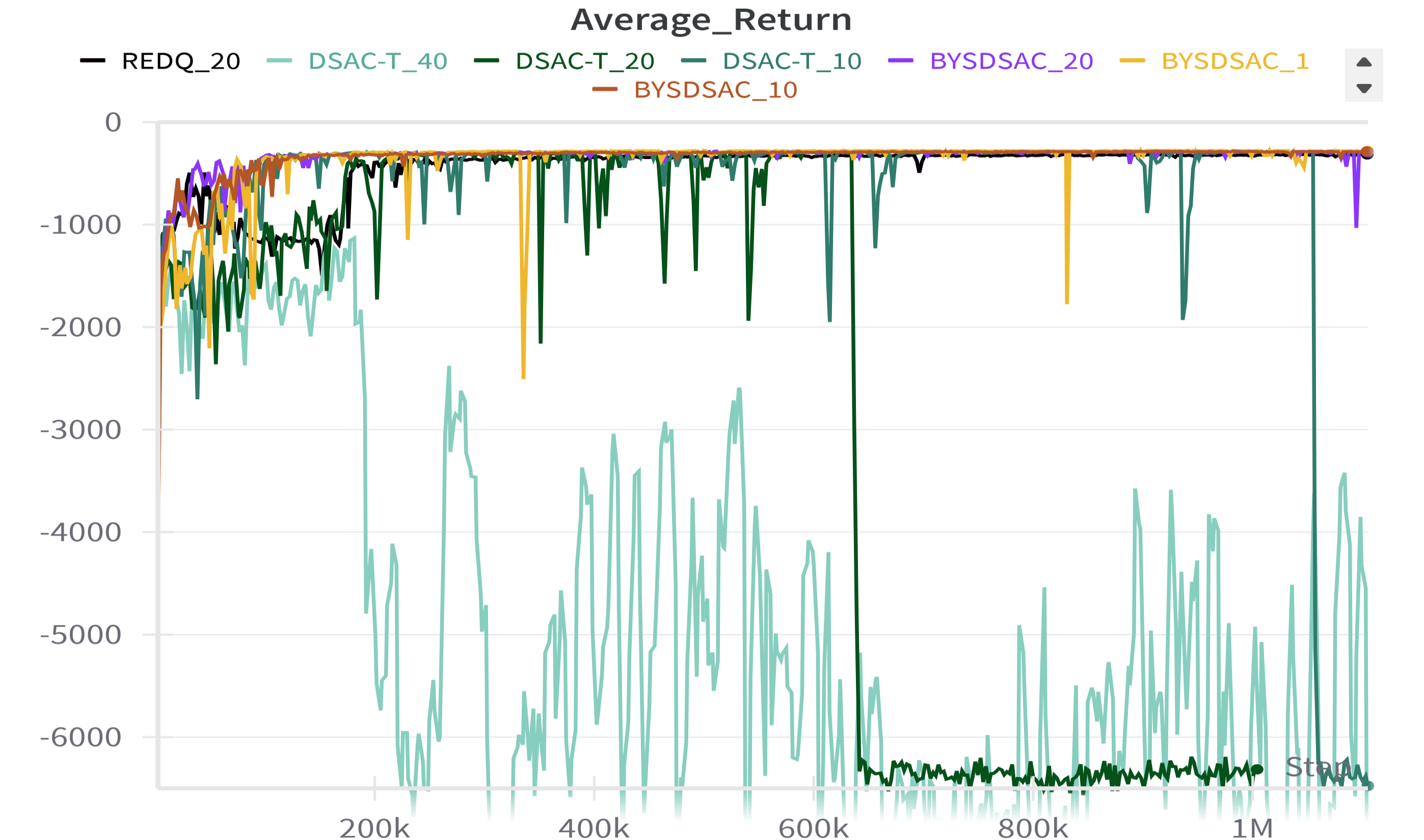}
  \caption{Comparison results of training with different replay ratios. The numbers following the type names represent the adopted replay ratios. From the results, it can be observed that, unlike DSAC-T which experiences training collapse with high replay ratios, Bayes-DSAC is able to utilize relatively high replay ratios for training.}
  \label{fig:result2}
\end{figure}

Additionally, we explored the replay ratio of the algorithm. Traditional methods usually cannot train multiple batches from a single environment interaction sample, leading to lower sampling efficiency. 
REDQ, due to randomly selecting a Q-function to compute the target return, can utilize a larger replay ratio. From the results in Fig~\ref{fig:result2}, it is evident that our proposed Bays-DSAC can choose a higher replay ratio than DSAC-T. When selecting a replay ratio of 20, the curve maintained training performance without performance degradation caused by overfitting, achieving results comparable to REDQ.

The experimental results verify the effectiveness of Bays-DSAC, showing that it improves the quality of value function estimation, resulting in faster algorithm convergence and higher environment sampling efficiency.

\subsection{Whole-body motion planning validation}
\begin{figure*}[htbp]
  \centering
  \subfloat[]
  {
  \includegraphics[scale=0.18]{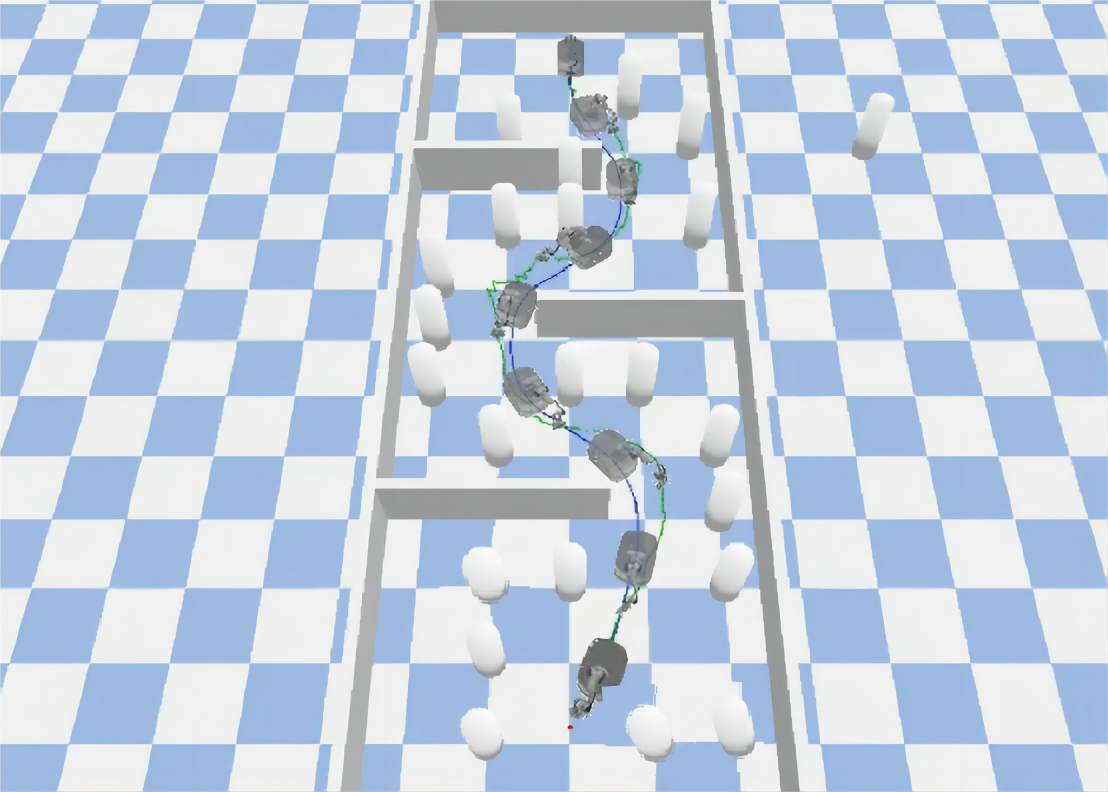}}
  \subfloat[]
  {
  \includegraphics[scale=0.18]{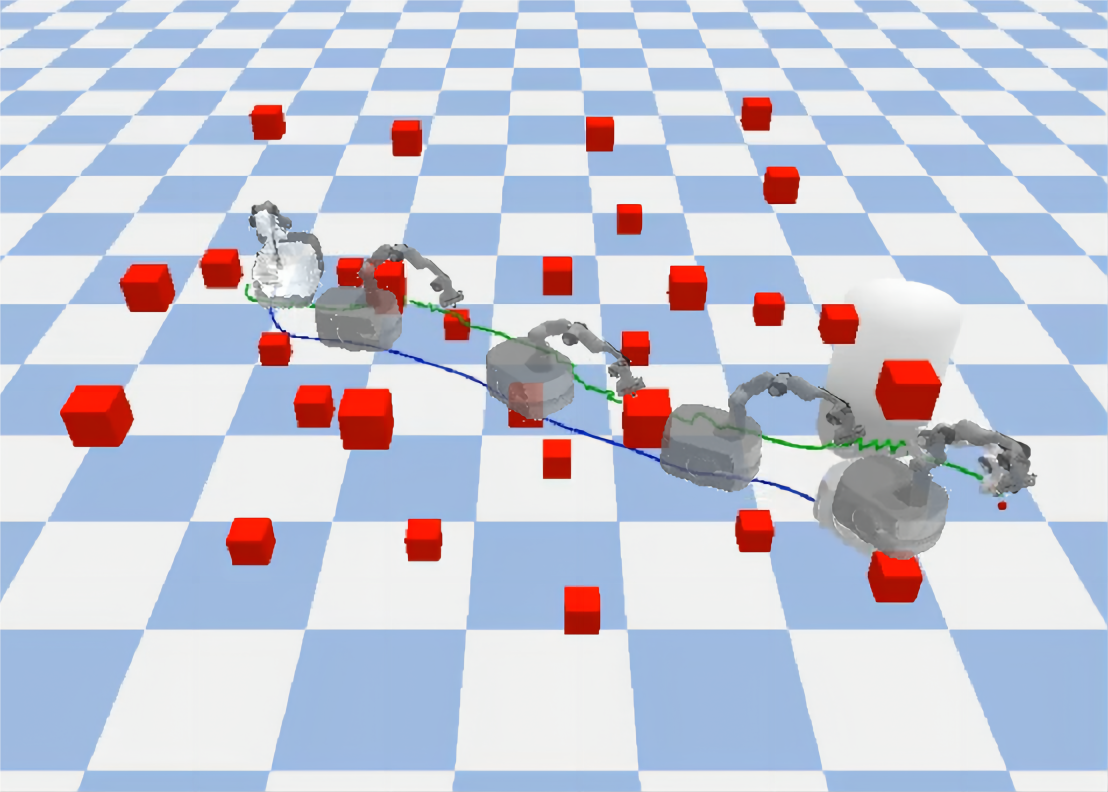}}
  \quad
  \subfloat[]
  {
  \includegraphics[scale=0.18]{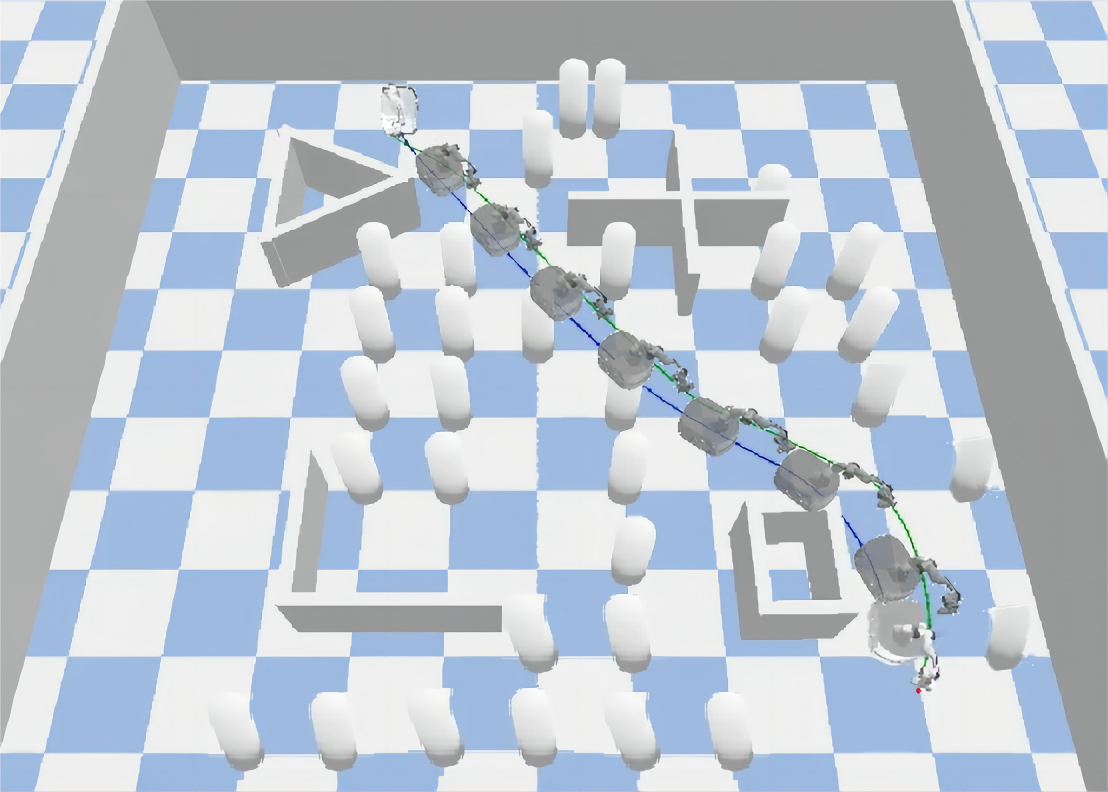}}
  \subfloat[]
  {
  \includegraphics[scale=0.18]{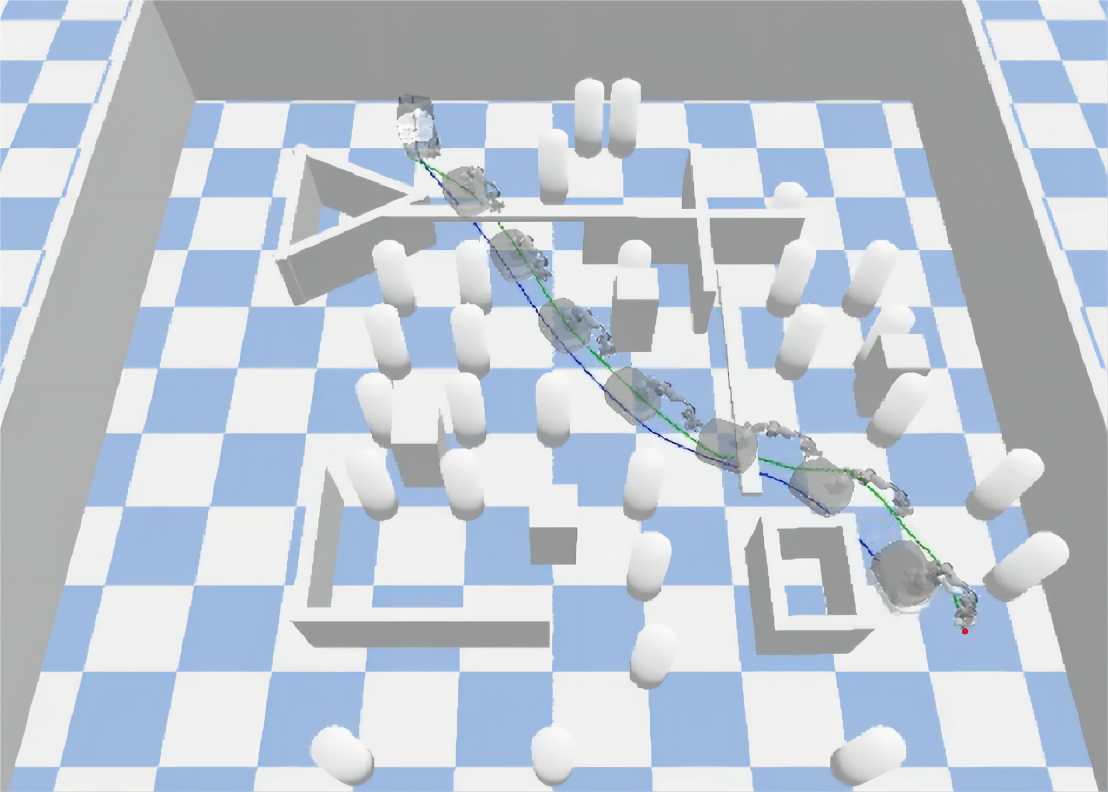}}
  \caption{
  Four cluttered scenarios(a-d) with the planned results by our hybrid whole-body motion planning framework.
  The red sphere in the figure shows the target EE position. The blue trajectory is the mobile base path, and the green is the end-effector trajectory.
}
  \label{fig:per9}
\end{figure*} 
In this section, the hybrid learning and optimization framework is tested and analyzed in simulation environments to verify the effectiveness and efficiency.
In all cases, the goal is to complete the reaching tasks quickly, efficiently and safely.
In the experiments, the mobile manipulator consists of a two-wheel differential-drive base Omeron and a 7-degrees-of-freedom Panda manipulator. The mobile base is equipped with an Intel RealSense D435 RGB-D camera. 
The simulation is implemented in Pybullet.

To evaluate the feasibility of the proposed framework(HORF), we select the Holistic reactive approach(HRA)\cite{Holistic}, reinforcement learning for mobil manipulator(RLMM)\cite{kindleWholeBodyControlMobile2020a}, perceptive MPC(P-MPC)\cite{pankertPerceptiveModelPredictive2020a} and dual-trajectory tracking methods(DTTM)\cite{wangReactiveMobileManipulation2024} as baselines.
Furthermore, we employ a hybrid framework combining SAC and SDF constraints(HFSS), as well as a hybrid framework integrating Bayes-DSAC and sampling-based obstacle avoidance constraints\cite{havilandNEONovelExpeditious2021a}(HFBS), to comparatively validate the efficacy of our proposed submodule.
Here we use SDF constraints to extend the original HRA. RLMM directly uses RL to learn velocities, and the reward function is consistent with \eqref{eq14}.
P-MPC is a predictive control strategy that extends MPC by introducing visual information. 
DTTM uses the A* and workspace search to construct constrained corridor and samples it to obtain feasible dual trajectories for the base and EE.
We randomly chose start and goal pairs in the four scenarios shown in Fig~\ref{fig:per9} to compare the five methods for arrival performance.
We employ the trajectory lengths(TL) of the mobile base and the EE, the Average velocity(AV), the total runtime(including planning and execution time), and the success rate to compare the performance of the methods. The results are shown in Table~\ref{table:1}.
When the planning is executed for more than 2 minutes or resulted in collisions, the task is considered a failure.

It can be observed from the results that the augmented HRA generally has longer trajectory lengths and a higher failure rate, because the HRA method tracks the direction of the target without adjusting the desired speed according to the environment. 
During the optimization process, trajectories in environments with many obstacles tend to exhibit more jitter and may get trapped.
The trajectory quality obtained by the RLMM is poorer. 
This is because RLMM directly learns joint velocities without imposing constraints and optimizations on other objective such as manipulability. 
P-MPC can handle cluttered scenes well and produce positive whole-body control results. 
However, the planning time is relatively long, and in some scenarios, it is prone to getting stuck in local optima, leading to planning failures.
The DTTM relies on a pre-found mobile base path when determining the EE motion region, but this process may require repetitive planning due to the lack of feasible paths, which extend the planning time.
Additionally, the dual trajectory tracking performance depends on the selection of hyperparameters. This results in suboptimal performance for DTTM in some environments.
In contrast, our proposed hybrid whole-body planning framework effectively combines the advantages of reinforcement learning in understanding the environment and the advantages of QP in constraining joint movements. It provides velocity recommendations based on the environment and uses SDF constraints and manipulability optimization to solve for reasonable and safe joint velocities. The reactive algorithm is fast in computation with a high planning success rate even in cluttered scenes
Moreover, HFSS, due to its SAC inferior training outcomes, ultimately yields suboptimal planning results, resulting in trajectories with increased jitter. The HFBS, which employs 40 sampled points for obstacle avoidance constraints, increases computational time and produces collisions due to precision issues in some environments. Comparative analysis reveals that our approach, utilizing Bayes-DSAC and robot-centric SDF constraints, demonstrates superior performance in whole-body motion planning task.
Experimental results show that our method can achieve safe, swift, and elegant control effects.

\begin{table*}[htbp]     
        \caption{The results of whole-body motion planning for mobile manipulator}  
        \centering                
        \vspace{0.03cm}      
        \label{table:1}               
        \renewcommand{\arraystretch}{1.0}
        \setlength{\tabcolsep}{4.4mm}{
        \begin{tabular}{@{}cccccccccl@{}}
        \toprule
        \qquad\textbf{Scene} &\textbf{Metric}  &\textbf{HORF}  &\textbf{HRA}  &\textbf{RLMM} &\textbf{P-MPC} &\textbf{DTTM} &\textbf{HFSS} &\textbf{HFBS}\\ 
        \midrule
        \qquad\multirow{6}{*}{Scene1} 
        & TL of MM(m)    & 12.826 & 14.702   & 15.925  & \textbf{12.131} & 12.759  & 13.523 & 12.895   \\
        & TL of EE(m)    & \textbf{15.608} & 17.791   & 21.352  & 15.724 & 15.783   & 16.267 & 15.651  \\
        & AV(m/s)        & 3.031       & 2.125   & \textbf{3.311}     & 3.098 &3.296 & 3.281 &3.034 \\
        & Total Runtime(s)& \textbf{5.750} & 6.940   & 6.549  & 6.276 & 5.589 & 6.038 & 6.810\\  
        & Success Rate(\%) & 95.87\% &65.15\%    &67.73\%   &\textbf{96.21\%} & 94.42\% &92.477\% &94.158\%\\ \midrule
        \qquad\multirow{6}{*}{Scene2} 
        & TL of MM(m)    & \textbf{6.301} & 9.215   & 10.248  & 8.012 & 7.702 & 7.144 & 6.509\\
        & TL of EE(m)    & \textbf{7.882} & 10.906   & 15.983  & 9.381 & 8.239    & 8.940 & 8.036\\
        & AV(m/s)        & \textbf{2.195} & 1.892   & 2.031  & 2.034 & 1.993  & 2.008 & 2.154\\
        & Total Runtime(s)& \textbf{3.990} & 5.841   & 7.969   & 5.812     & 6.139 & 4.182     & 5.403\\ 
        & Success Rate(\%) & \textbf{98.97\%} & 70.12\%   & 75.54\%  &81.82\% & 92.67\% & 92.883 & 96.608\\ \midrule
        \qquad\multirow{6}{*}{Scene3} 
        & TL of MM(m)    & \textbf{9.347} & 12.052   & 11.981  &10.094  &10.129    & 10.437 & 9.804\\
        & TL of EE(m)    & \textbf{9.661} & 13.671   & 18.762  &10.952 & 11.001    &10.866 & 9.961\\
        & AV(m/s)        & \textbf{2.821} & 2.191   & 2.479  & 2.702 & 2.814  & 2.744 & 2.803\\
        & Total Runtime(s)& \textbf{3.794} & 6.679   & 7.668  & 5.453    & 4.509  & 4.270    & 4.740\\ 
        & Success Rate(\%) & \textbf{98.91\%} & 71.07\%   & 81.05\%  &98.24\% &97.82\% &96.288\% &97.814\%\\ \midrule
        \qquad\multirow{6}{*}{Scene4} 
        & TL of MM(m)    & \textbf{9.498} & 13.087   & 12.032  &10.512 & 10.046   &10.517 & 10.095\\
        & TL of EE(m)    & \textbf{10.088} & 18.824   & 21.761  &12.004 & 11.893   &11.338 & 10.854 \\
        & AV(m/s)        & \textbf{2.812} & 2.125   & 2.321  &2.703 & 2.781  &2.717 & 2.794\\
        & Total Runtime(s)& \textbf{3.907} & 9.058   & 9.575  & 6.840 & 7.476 & 4.702 & 5.374\\ 
        & Success Rate(\%) & \textbf{97.32\%} & 48.82\%   & 53.31\%  &91.85\% & 95.94\% &94.803\% & 96.367\%\\
        \bottomrule
        \end{tabular}}
        \end{table*}

\section{ Conclusion}
In this article, we propose a hybrid learning and optimization framework for reactive whole-body motion planning of mobile manipulators in cluttered environments.
To enhance the quality of value function estimation, we propose Bayes-DSAC, an off-policy reinforcement learning method. 
Additionally, we introduce a SDF constraints QP to enhance the system's joint obstacle avoidance capabilities. 
Many experiments have verified that our framework generates efficient and safe motion planning results compared to other methods, and improves the obstacle-avoidance motion capability of mobile manipulators in complex environments.

However, the current system is limited by the finite camera perception range, and there are safety concerns. In future work, we will study the collaboration between the gaze direction of camera and the mobile manipulators whole-body planning to  enhance the operational safety.

\bibliographystyle{IEEEtran}
\bibliography{references}


\end{document}